\documentclass[runningheads,a4paper]{llncs}
\usepackage[T1]{fontenc}
\usepackage{amssymb}
\usepackage{graphicx}
\usepackage{amsmath}
\usepackage{multirow}
\usepackage{booktabs}
\usepackage[caption=false]{subfig}
\newcommand{\ra}[1]{\renewcommand{\arraystretch}{#1}}

\usepackage{url}
\newcommand{\keywords}[1]{\par\addvspace\baselineskip
\noindent\keywordname\enspace\ignorespaces#1}

\usepackage[table]{xcolor}

\begin{document}

\mainmatter  

\title{Racing Multi-Objective Selection Probabilities}

\titlerunning{Racing Selection Probabilities}

%
%
\author{G. Marceau-Caron\inst{1,2} \and M. Schoenauer\inst{2}}
%
\authorrunning{Ga\'etan Marceau-Caron et al.} 
%
\tocauthor{Ga\'etan Marceau-Caron, Marc Schoenauer}
\institute{Thales Air Systems, Rungis, France\\
\and
TAO Project, INRIA Saclay \&  LRI Paris-Sud University, Orsay, France\\
\email{\{marc.schoenauer,gaetan.marceau-caron\}@inria.fr}\\ 
}

%
%

\toctitle{Lecture Notes in Computer Science}
\tocauthor{Authors' Instructions}
\maketitle

\begin{abstract}
In the context of Noisy Multi-Objective Optimization, dealing with uncertainties requires the decision maker to define some preferences about how to handle them, through some statistics (e.g., mean, median) to be used to evaluate the qualities of the solutions, and define the corresponding Pareto set. Approximating these statistics requires repeated samplings of the population, drastically increasing the overall computational cost. To tackle this issue, this paper proposes to directly estimate the probability of each individual to be selected, using some Hoeffding races to dynamically assign the estimation budget during the selection step. The proposed racing approach is validated against static budget approaches with NSGA-II on noisy versions of the ZDT benchmark functions.

\end{abstract}

\section{Introduction}
Uncertainty handling is an important aspect of optimization since it concerns most, if not all, real-world applications. 
Optimizing uncertain objectives aims at taking into account modeling inaccuracies, measurement errors from sensors, or prediction errors, that will interfere with the beliefs of a decision maker about the environment. 
Therefore, optimization under uncertainty must include some mechanisms that ensure, one way or another, that the proposed solutions are effective, according to the user's point of view w.r.t. optimality. 
And whereas several definitions of such effectiveness can occur in the simplest case of a single objective, the complexity of optimizing multiple uncertain objectives increases drastically with the number of objectives.

The general framework of this work is that of multi-objective optimization in uncertain context. The degrees of freedom of the decision maker are the variables of the optimization problem, defined on the decision space, and observables are some responses of the system when setting these variables. However, the same setting will result in different responses every time it is used, and the output of the system thus defines a probability distribution, conditionally dependent on the decision variables.
%
%
The goal of the optimization process is then to find the values of the decision variables that will optimize some statistics on this probability distribution. The choice of these statistics depends on the user's goal and preferences. The average or the median are common choices, though probably sometimes only because of the lack of efficient methods to handle other statistics. For instance, risk-adverse users will prefer to minimize the consequences of the worst outcomes, while risk-affine users will maximize their possible ``profit'' even if it comes at high risk, optimizing the {\em value at risk} for a given risk level.

Except when the type of noise is known --a totally unrealistic hypothesis-- a common way to compute the desired statistics is to sample the fitness of each individual as many times as necessary to obtain a good estimation thereof, and the amount of computation per individual is user-defined, uniformly over the individuals and the generations. In the single-objective framework, an alternative has been proposed, using the idea of {\em races} \cite{Heidrich-Igel-ICML09}, minimizing the number of re-evaluations while keeping a high confidence level on the results. But only limited attempts have been made in the multi-objective framework (see Section \ref{State-of-the-art}). 

The approach proposed in this paper, {\em Racing Selection Probability} (RSP), is an attempt, in the multi-objective case, to dynamically decrease the number of sampling of all individuals by applying the principles of Hoeffding races directly on the estimation of the probability of being selected for an individual, using bounds on the behavior of that probability to decide as early as possible when to definitely select or discard an individual, for a given confidence level. Bounds on any statistics can be used, and straightforwardly embedded in any unmodified EMOA, thus allowing to handle any preference of the user. Furthermore, any type of noise can be handled that way.

The paper is organized the following way. Section \ref{State-of-the-art} briefly surveys state-of-the-art methods for uncertainty handling in Evolutionary Multi-Objective Optimization. Section \ref{racing} introduces the Hoeffding's inequality used in RSP. Section \ref{results} presents experimental results on perturbed ZDT test functions (with different types of noise) where RSP is compared to the two basic noise-handling methods, the implicit and static averaging.
More details and results are available in the corresponding Technical Report\footnote{http://hal.inria.fr/hal-01002854/en}.

\section{Uncertainty Handling in Multi-Objective Evolutionary Optimization}
\label{State-of-the-art}

The context of this work is that of Multi-Objective Optimization with Uncertainty. On the space of decision variables $\cal X$, several conflicting objectives $f_1, \ldots, f_k$ are defined (to be minimized, w.l.o.g.), and, as discussed above, the outcome of any given setting of the decision variables is a probability distribution $\boldsymbol f$ over the objective space $\cal F \subset {\mathbb R}^{\mbox{$k$}}$ that depends on the values of the variables and on some additional unknown external random variable $\boldsymbol \varepsilon$, aka noise. In particular \cite{basseurZitzler:EvoApps2006}, there is no ``true'' value of the objectives to which some random noise is added. 
Formally \cite{trautmanNaujoks:CEC2009,vossIgel:PPSN2010}, the Multi-Objective Noisy Optimization Problem (MNOP) can be written as 
\begin{equation}
\label{MNOP}
 \min_{\mathbf{x} \in {\cal X}} ( (\mathbf{f}|\mathbf{x}, {\boldsymbol \varepsilon}) = (f_1, \ldots, f_k|\mathbf{x}, {\boldsymbol \varepsilon}))
 \end{equation}

where $\mathbf{f}$ is a random variable taking values in $\cal F$, and each $f_i$ is a real-valued random variable, a coordinate of $\mathbf{f}$.

Even in the single objective case, the minimization of a random variable does not make much sense. So the user must complete the problem definition by providing some preferences through some statistics over that random variable (e.g., minimizing the mean, the median, the 5\% percentile, the variance with constraint on the mean, \ldots). The situation is the same in the multi-objective case, except that there doesn't exist any total order on the samples of the random variable of interest. In the deterministic case, Pareto dominance has proved useful, and the notion of Pareto front is accepted as a way to describe interesting solutions of the multi-objective problem at hand. In particular, several multi-objective optimization algorithms have been proposed, among which Evolutionary Multi-Objective Algorithms (EMOAs) (see e.g., \cite{surveyZhou2011}).
And because uncertainty is ubiquitous in real-world problems, MNOPs have also been well studied, though not always with such a degree of generality. 

\subsection{Previous Work}
\label{sec:previousWork}
A first approach is to port to multi-objective context the single-objective {\em static averaging} techniques, that re-evaluate every individual $N$ times at each generation (also called {\em implicit averaging} if $N=1$). 

Several works consider the specific case of additive noise of known type: the random variable $(\boldsymbol f|\mathbf{x}, {\boldsymbol \varepsilon})$ is of the form $\mathbf g(x) + {\boldsymbol \epsilon}$ for some function $\mathbf g(x)$ and some partly known noise ${\boldsymbol \epsilon}$. Depending on the form of ${\boldsymbol \epsilon}$, approximation of the probabilistic dominance (probability that an individual Pareto-dominates another one) can sometimes be computed at low computational cost. In \cite{Teich:EMO2001}, the noise is supposed bounded, and exact calculations are done for uniform noise; In \cite{Hughes:EMO2001}, the noise is assumed Gaussian with known variance (that can be computed off-line from static samples).
This work is extended in \cite{Fieldsend:CEC2005} to the case of unknown (and non-uniform) variance. Later, \cite{Eskandari:CEC07} proposed another way to compute the probability with more general hypotheses, but going back to using a fixed number of samples (15 in experiments).  In any case, it is clear that the hypothesis of a known type of noise is highly unrealistic in practice.

An approach that is specific to indicator-based algorithms is proposed in \cite{basseurZitzler:EvoApps2006}, that does not make any hypothesis about the noise and uses the general model of Equation \ref{MNOP}: the indicator $\varepsilon^+$ is approximated using averages (over 5 samples), and is used within the environmental selection. However, the problem being solved there is the minimization of the expectation of the indicator at hand (w.r.t. some reference set), that cannot be adapted to the user's preferences.

Several works propose different approaches to probabilistic dominance for the general MNOP (equation \ref{MNOP}).
Pareto Dominance in Uncertain environments(PDU) \cite{trautmanNaujoks:CEC2009} uses the convex hull of a fixed number of samples (10 in the paper) to estimate both the mean and its uncertainty. In \cite{vossIgel:PPSN2010}, PDU evaluates the certainty of the mean using quartiles on each dimension, and some races are run for each objective, from \cite{Heidrich-Igel-ICML09}, with confidence 0.0001 and maximum race length 15. This latter work however assumes that the noise distribution is symmetrical, and Suzuki and collaborators propose another Pareto Dominance operator that does not need that hypothesis \cite{BoonmaSuzuki:ICTAI2009} using a CPU-expensive SVM construct over the samples; \cite{Phan:2012} improves the method using a non-parametric Mann-Whitney U-test. However, both works use a fixed number of samples (resp. 30 and 20) to estimate the dominance operator.

In \cite{masterBranke2009}, six different resampling approaches are compared. All but one use some absolute criteria that only depend on some statistics on the previous samples and the individual at hand to decide on early stop of the resampling procedure and derive an estimation of the mean of the sample with known confidence. That mean is then used as the fitness in a standard EMOA. The last procedure (termed OCBA) is the closest to RSP proposed here, in that it makes the minimal global sampling allocation to estimate the confidence in a partition of the population into a non-dominated and a dominated sets. However, the calculation of the confidence assumes Gaussian noise on all objectives.

\subsection{Discussion, and Rationale for RSP}
Our goal is to design, within a given EMOA, an approach that will limit the number of resampling while preserving some confidence on the resulting Pareto-based selection, for a wide range of statistics describing the user's preferences, and without any requirement on the type of noise. Most of the works listed above, however, use a fixed user-define resampling budget (except \cite{masterBranke2009} and \cite{vossIgel:PPSN2010}). Furthermore, either they derive estimations of the mean of a sample with some confidence interval -- and this does not allow to derive confidence bounds on the comparison between those means (except in specific cases, e.g. Gaussian distributions); or they do derive probabilistic Pareto dominance, with known confidence, but omit the second component of Pareto-based selection, the diversity preserving mechanism (the case of indicator estimation \cite{basseurZitzler:EvoApps2006} is different, but strictly limited to indicator-based EMOAs).

The idea of RSP borrows from \cite{Heidrich-Igel-ICML09}, like \cite{vossIgel:PPSN2010} cited above, is using Hoeffding races\footnote{\cite{Heidrich-Igel-ICML09} also advocates Bernstein races when the range of values is not know -- which is not the case here. Hence Bernstein races will not be mentioned here.} to decrease the number of resampling while nevertheless guaranteeing some level of confidence on the statistic at hand. But contrary to the works above (including \cite{vossIgel:PPSN2010}), {\em Racing Selection Probability}, as its name claims, will perform the race on the probability of an individual to be selected by the selection mechanism of the chosen EMOA.

\section{Racing Selection Probability}
\label{racing}
Let us assume some selection procedure in an existing MOEA (e.g., non-dominated sorting + crowding distance for NSGA-II \cite{DebBook2001}) that aims at selecting $\mu$ individuals out of a population of size $\lambda$. The basic idea of RSP, inspired by \cite{Heidrich-Igel-ICML09}, is to estimate, for any individual $i$, with as few samples of fitnesses as possible thanks to Hoeffding bounds, the probability $p^{sel}_i$ that $i$ will be selected. 



Hoeffding's inequality states that, for any random variable $X$ with range width $R$, and for any confidence level $1-\delta$, the absolute difference between the expectation and the empirical mean computed using $t$ samples is upper-bounded by $R \sqrt{\log(2/\delta)/2t}$.

Every time all $\lambda$ individuals are resampled, the standard selection procedure of the EMOA at hand is applied to the current sample, determining the selected $\mu$ ones. This results in a new sample for probabilities $p^{sel}_i$. For any $\delta$, lower and upper values for all $p^{sel}_i$ can be computed at confidence level $1-\delta$, thanks to Hoeffding's bound applied to $p^{sel}_i$. Any  individual $i$ whose lower bound for $p^{sel}_i$ is larger than the upper bound of at least $\lambda - \mu$ other $p^{sel}_k$ is definitely selected and leaves the race. Symmetrically, any  individual $i$ whose upper bound for $p^{sel}_i$ is smaller than the lower bound of at least $\mu$ other $p^{sel}_k$ is definitely discarded and leaves the race. Remain in the race the uncertain individuals, and only those are resampled again at next iteration. The race ends when either $\mu$ individuals are definitely selected, or $\lambda-\mu$ individuals are definitely discarded, or some maximum number of resampling $T_{Max}$ have been 
done. 
In the latter case, selection is made without Hoeffding guarantee. 
Nevertheless, the proposed procedure allows to quickly select the most promising individuals with given confidence. 
Note that each selection step can also be done on some statistic for each individual given the past $t$ samples. This will be illustrated in Section \ref{results} where variants using the average or the median will be used, instead of the most recent sample. These variants will be termed $RSP_{AVG}$ and $RSP_{MED}$ respectively, the variant that does not use any statistic being denoted by $RSP_{I}$.

There are however some specificities to the multi-objective context. First, the selection step usually involves the whole population, be it indicator based, or using some diversity secondary criterion. Hence the individuals that have left the race should nevertheless be taken into account for the next selection steps - but without being themselves re-evaluated, of course. A bootstrap procedure is used here, to mimic an ever growing sample without any resampling. Note that bootstrap is also used to avoid resampling individuals that have not been modified by the variation operators from one generation to the next.

Finally, it might be beneficial to detect early that some race will not end before the maximum number of samples because of actual ties between individuals that remain in the uncertain set. Here, when the sum of absolute pairwise differences of the empirical mean of the  $p^{sel}_i$ becomes lower than a given threshold called {\em Proximity Threshold}, the race stops and the $\mu$ best individuals according to the current selection policy are returned.

\section{Experimental Results}
\label{results}

\subsection{Experimental Conditions}
Five methods have been experimentally compared: the implicit averaging, and two variants of the static sampling, whether the average or the median of the samples is used for the selection (see Section \ref{sec:previousWork}); and 3 variants of RSP, whether the last sample, the average or the median of the previous samples are used in the selection (see Section \ref{racing}). All RSP variants have been implemented within NSGA-II with standard SBX crossover and polynomial mutation.
A common parameter of static sampling and RSP is the {\em Sampling Budget}, that will denote the fixed number of samples for each individual in the static case, and the maximum length of the races in RSP. RSP also requires a {\em Confidence Level} and the {\em Proximity Threshold} (see previous Section \ref{racing}). 

The testbench is based on the classical ZDT suite, used either as is (deterministic setting), or with known additional noise: Gaussian noise, that should favor the average estimator compared to the median estimator, the empirical average being the minimum-variance unbiased estimator of the expectation of a normal distribution with unknown mean and variance; Cauchy noise, that has an infinite mean, hence the mean estimator should be perturbed because of the outliers; and Gumbel noise, an asymmetrical distribution with finite moments that is used in extreme value theory to simulate rare events (its location parameter is chosen in order to center the median). 

The goal of the experiments is to study the impact of the two parameters {\em Sampling Budget} and {\em Confidence Level}, and possibly their interaction, e.g., if the required confidence level is too high, all races will reach the maximum budget, and RSP amounts to static sampling.
All parameter values (for the algorithms and the noise models) that have been used for these experiments are listed in Table \ref{table:exp_conds}). All runs were limited to 100k evaluations, except ZDT6 (500k), and 25 independent runs were run for each parameter setting. Due to space limitation, results on ZDT2 are not shown, but are quite similar to those of ZDT4.

{\small
\begin{table*}[tb]
\centering
\begin{tabular}{@{}lllll@{}} \toprule
  ZDT Functions:& \{1,2,3,4,6\} & \phantom{abc} & Number of runs: & 25   \\
  Deterministic(DE): & Dirac Delta Function & \phantom{abc} & Gaussian noise(GA): & $0.25 * \mathcal{N}(0,\mathbb{I})$ \\
  Cauchy Noise(CA): & $(0,0.25)$ & \phantom{abc} & Gumbel noise(GU): & $(2, 2 \ln(\ln(2))$  \\
\hline
  Population Size: & 100 & \phantom{abc}& Nb Eval.: & \{100k,500k\} \\ 
  SBX Crossover: &$p_c = 1.0$, $\eta=20$ & \phantom{abc}& Polynomial Mut.: & $p_m=1/|\mathcal{X}|$, $\eta=20$ \\ 
\hline
  Confidence Level: & $\{0.25,0.95\}$ & \phantom{abc}& Proximity Thres.: & 0.5  \\ 
  Sampling Budget: & $\{5,10,15,20,30,50\}$ & \phantom{abc}& \multicolumn{2}{l}{Estimators: \{None, AVeraGe, MEDian\}}\\
  \bottomrule
\end{tabular}
\caption{\label{table:exp_conds}Parameters for (top to bottom) benchmarks and noise; static sampling; RSP}
\end{table*}
}


\subsection{Results}
\label{details-results}
The performances are compared using the {\it difference hypervolume} indicator w.r.t. the real Pareto front, on the normalized objective space. 
The normalization is done with respect to the Nadir point, computed from the union of the exact Pareto front and every point generated by each algorithm for a given function and a given noise.
Statistical significance is attested by p-values of the Wilcoxon signed-rank test.
Pisa performance assessment tools (\url{http://www.tik.ee.ethz.ch/sop/pisa}) was used to compute the hypervolumes.

Each plot of the following figures summarizes the results obtained by all algorithms on one function with one type of noise (or no noise at all): each plot displays several boxplots, each boxplot represents the statistics of the 25 hypervolume values at the end of each of the 25 runs for the corresponding setting. 
Each plot is divided into six regions.
First boxplot is that of the implicit averaging I. 
Next 2 regions give the results of the static sampling (resp. $AVG$ and $MED$), and display 6 boxplots each, corresponding to the 6 {\em Sampling Budget} values of Table \ref{table:exp_conds}.
Next 3 regions give the results for $RSP_I$, $RSP_{AVG}$ and $RSP_{MED}$ resp. For each region, there are 6 subregions (the 6 values of {\em Sampling Budget}) with two boxplots each, one for each  confidence level (25\%, 95\%). 

\begin{figure}[tb]
\hskip 0cm
\parbox{\textwidth}{
    \includegraphics[width=0.5\textwidth,height=0.15\textheight]{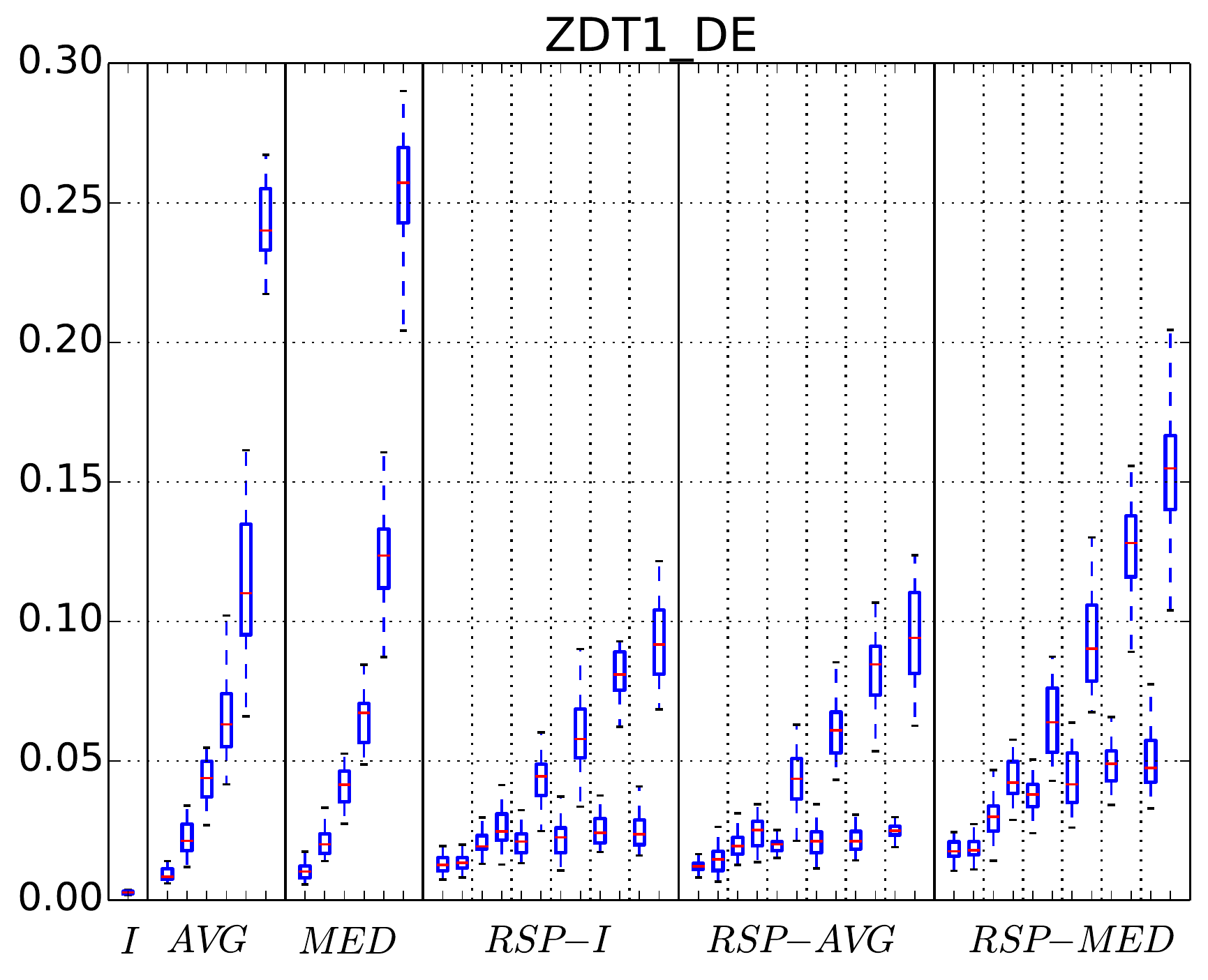} 
\hskip 0cm
    \includegraphics[width=0.5\textwidth,height=0.15\textheight]{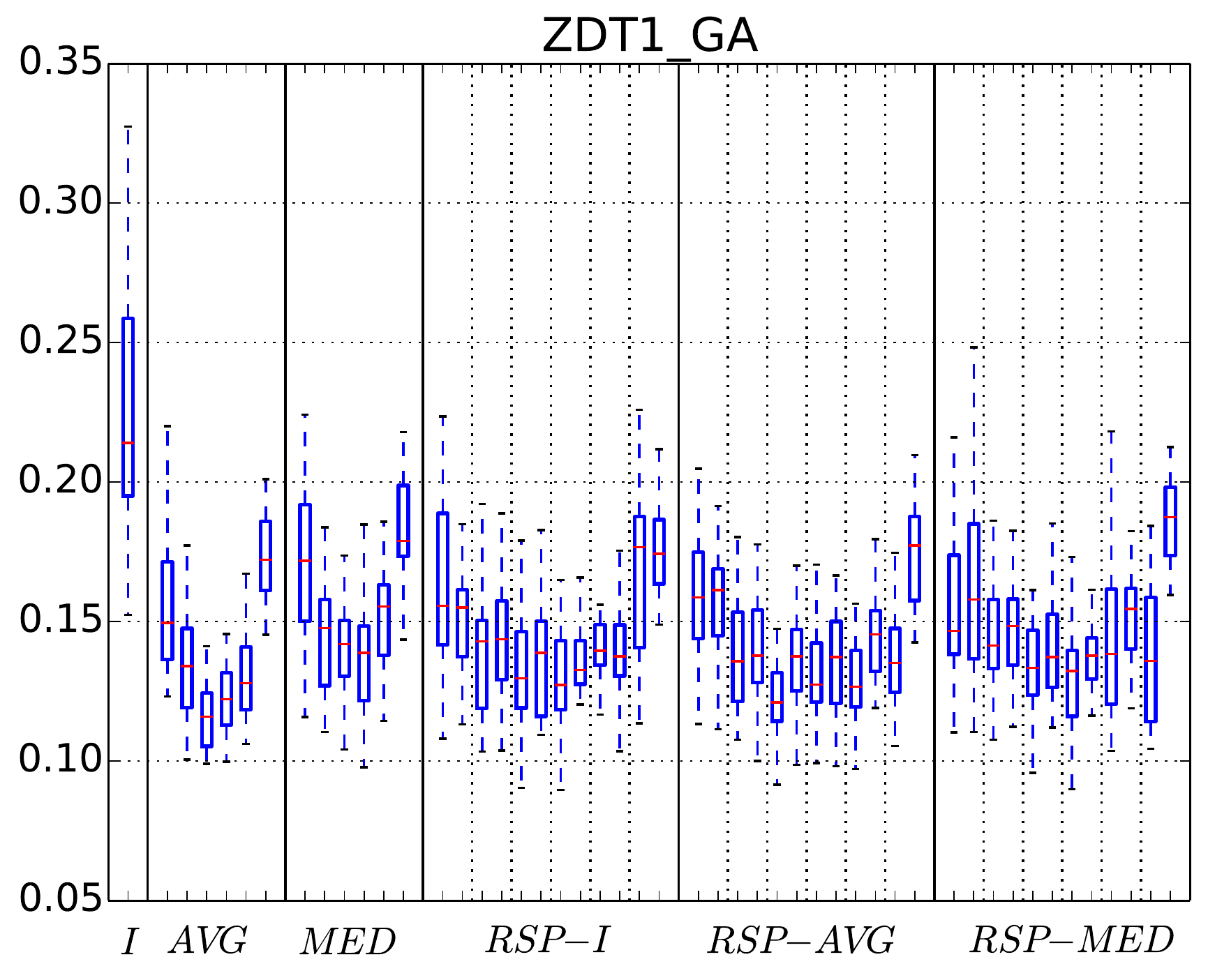}\\
    \includegraphics[width=0.5\textwidth,height=0.15\textheight]{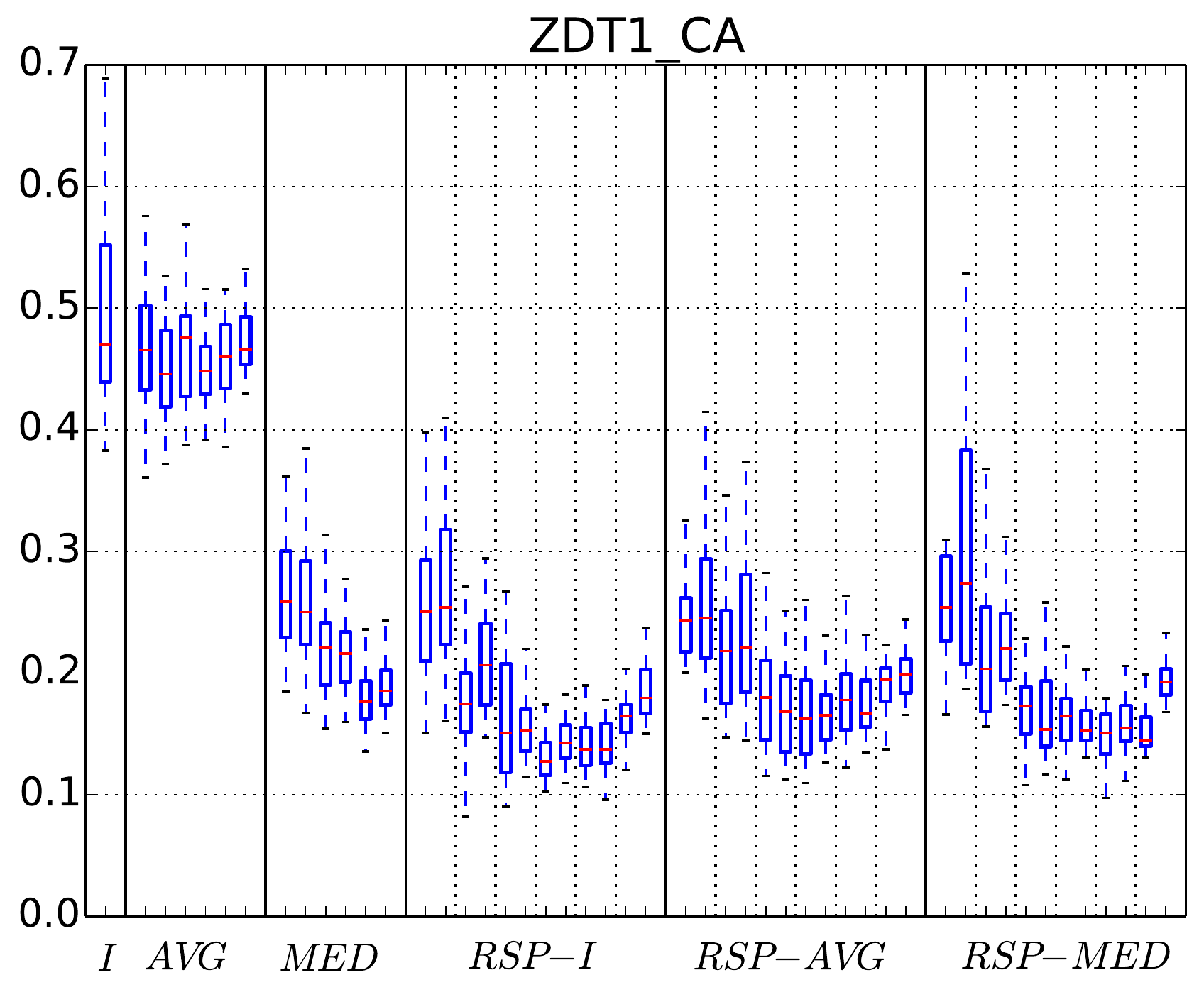}
\hskip 0cm
    \includegraphics[width=0.5\textwidth,height=0.15\textheight]{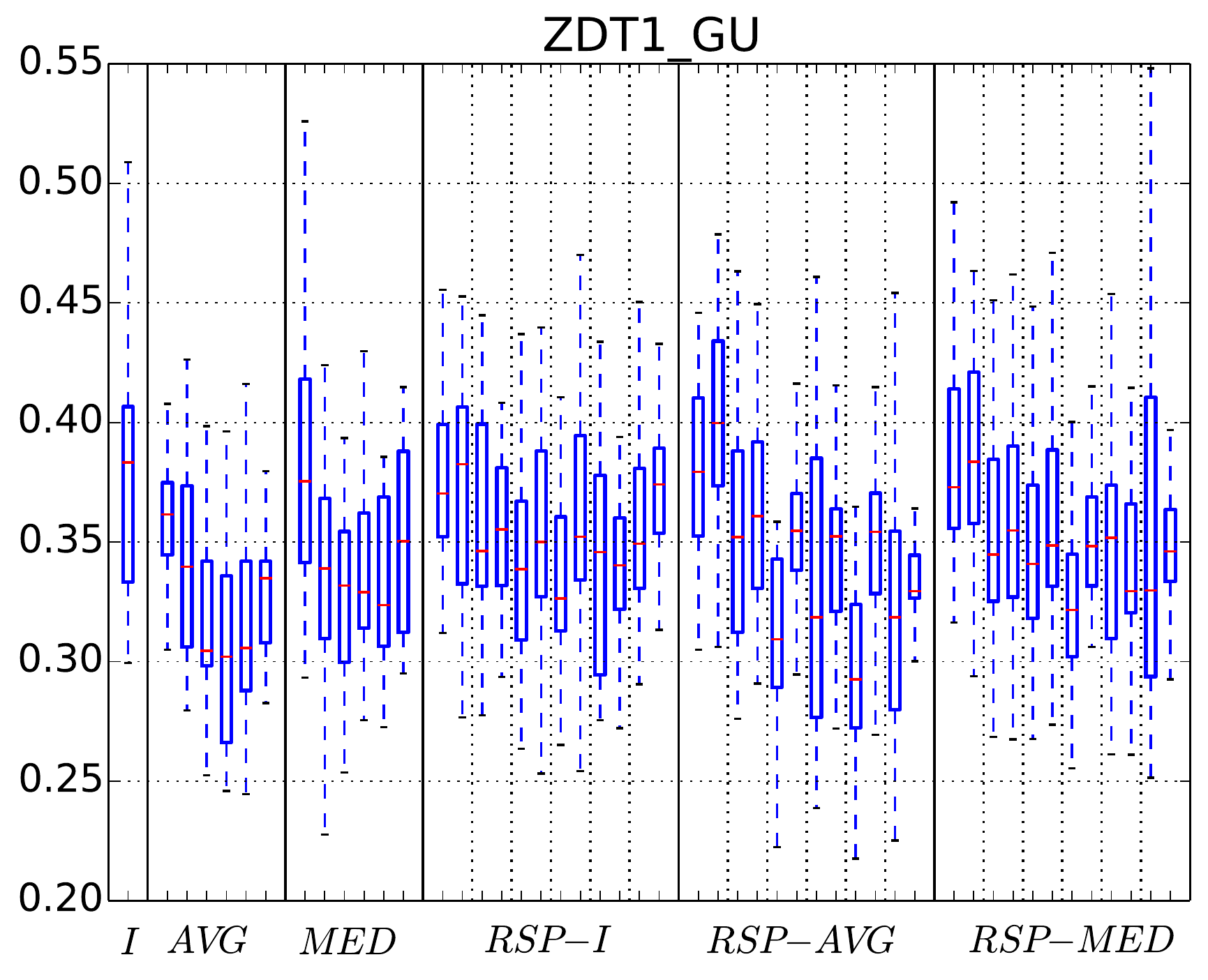}\\
    \includegraphics[width=0.5\textwidth,height=0.15\textheight]{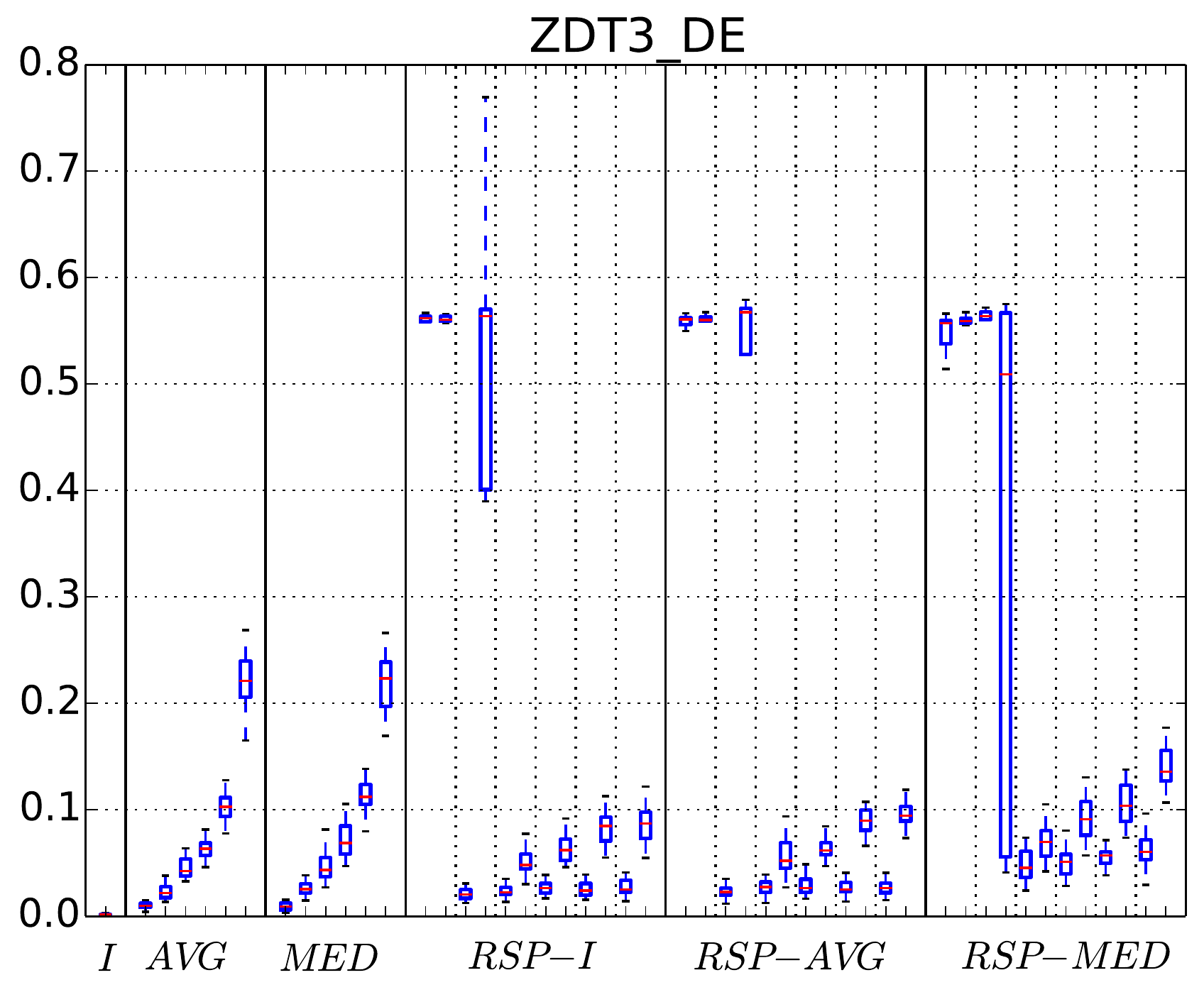}
\hskip 0cm
    \includegraphics[width=0.5\textwidth,height=0.15\textheight]{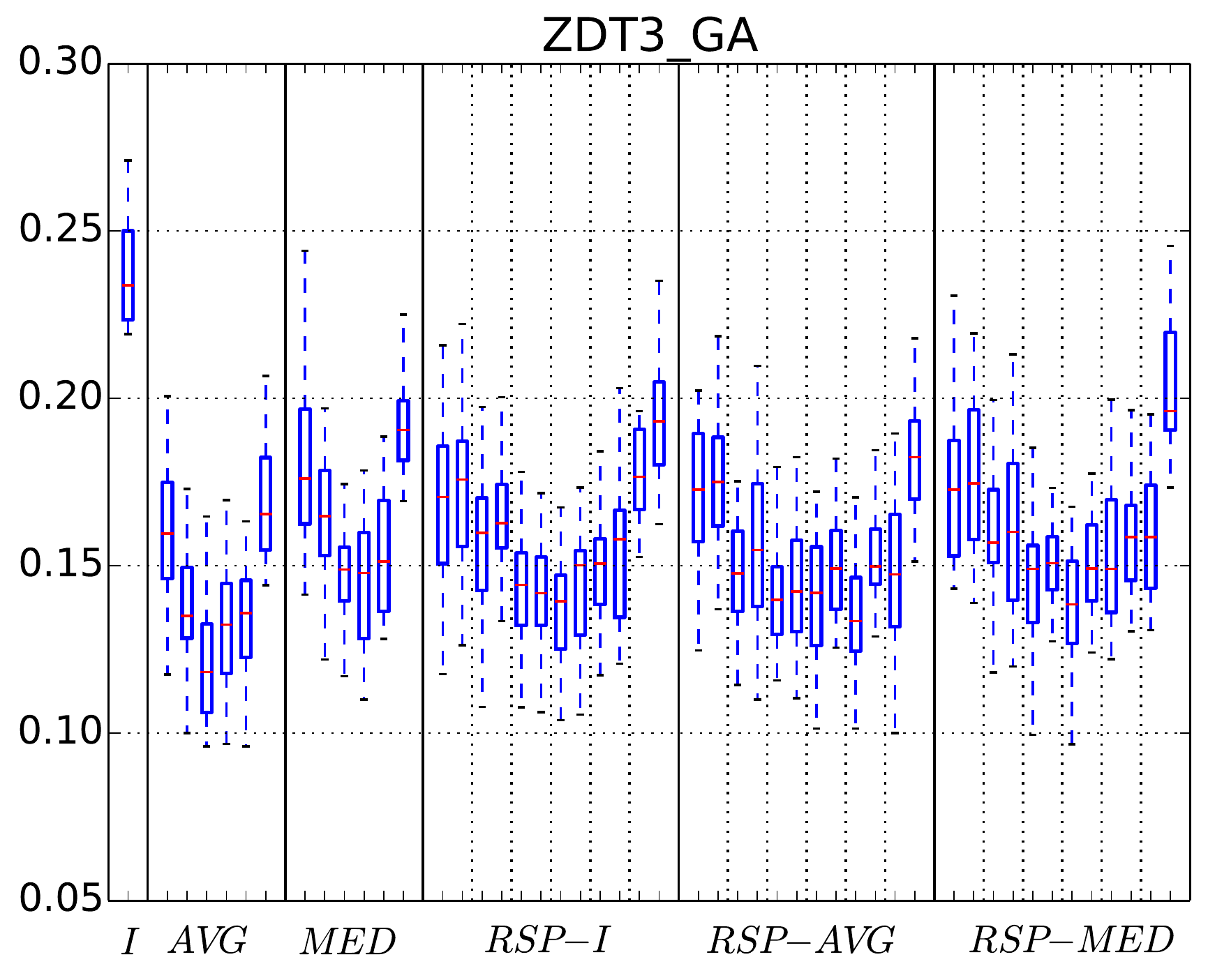}\\
    \includegraphics[width=0.5\textwidth,height=0.15\textheight]{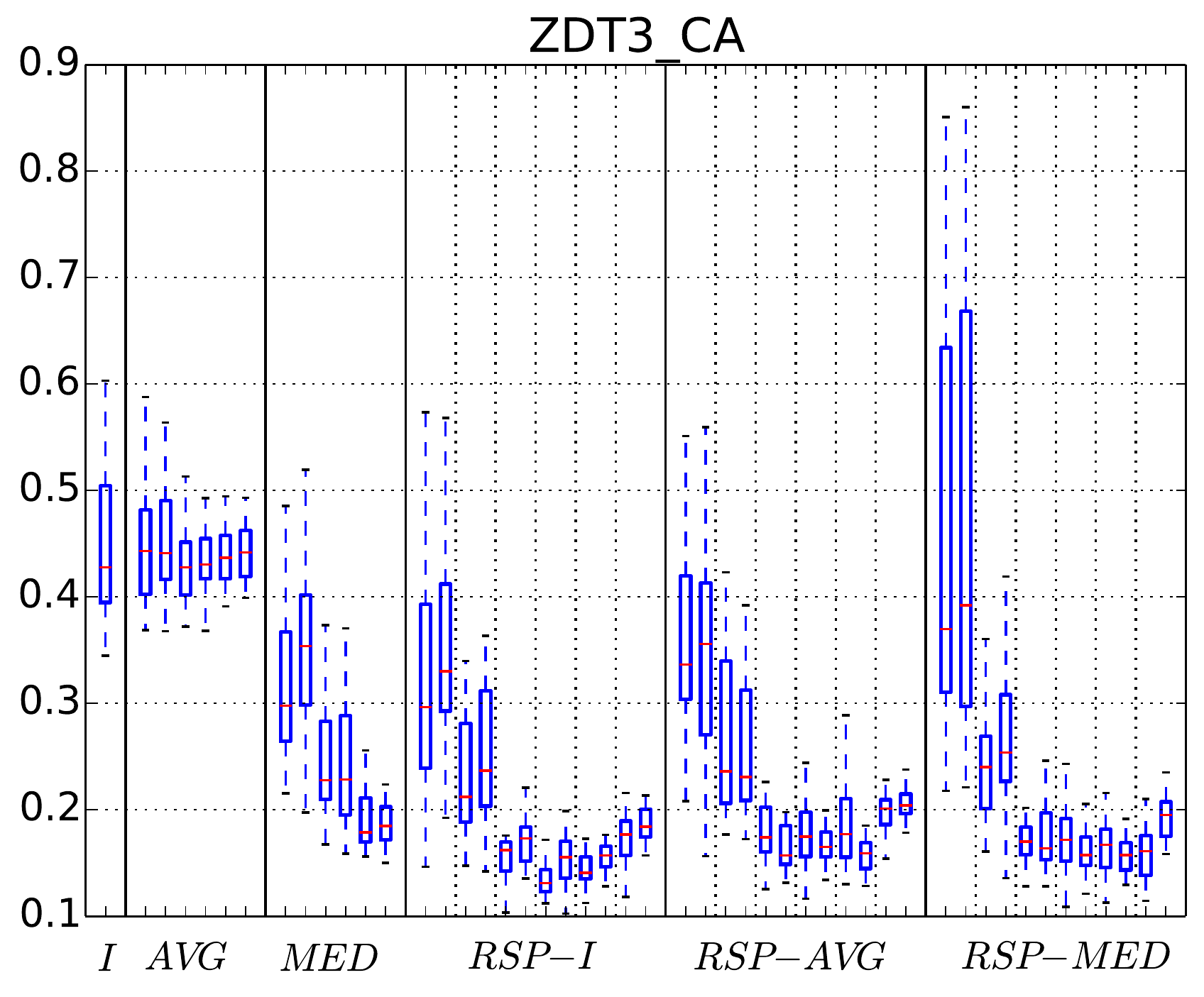}
\hskip 0cm
    \includegraphics[width=0.5\textwidth,height=0.15\textheight]{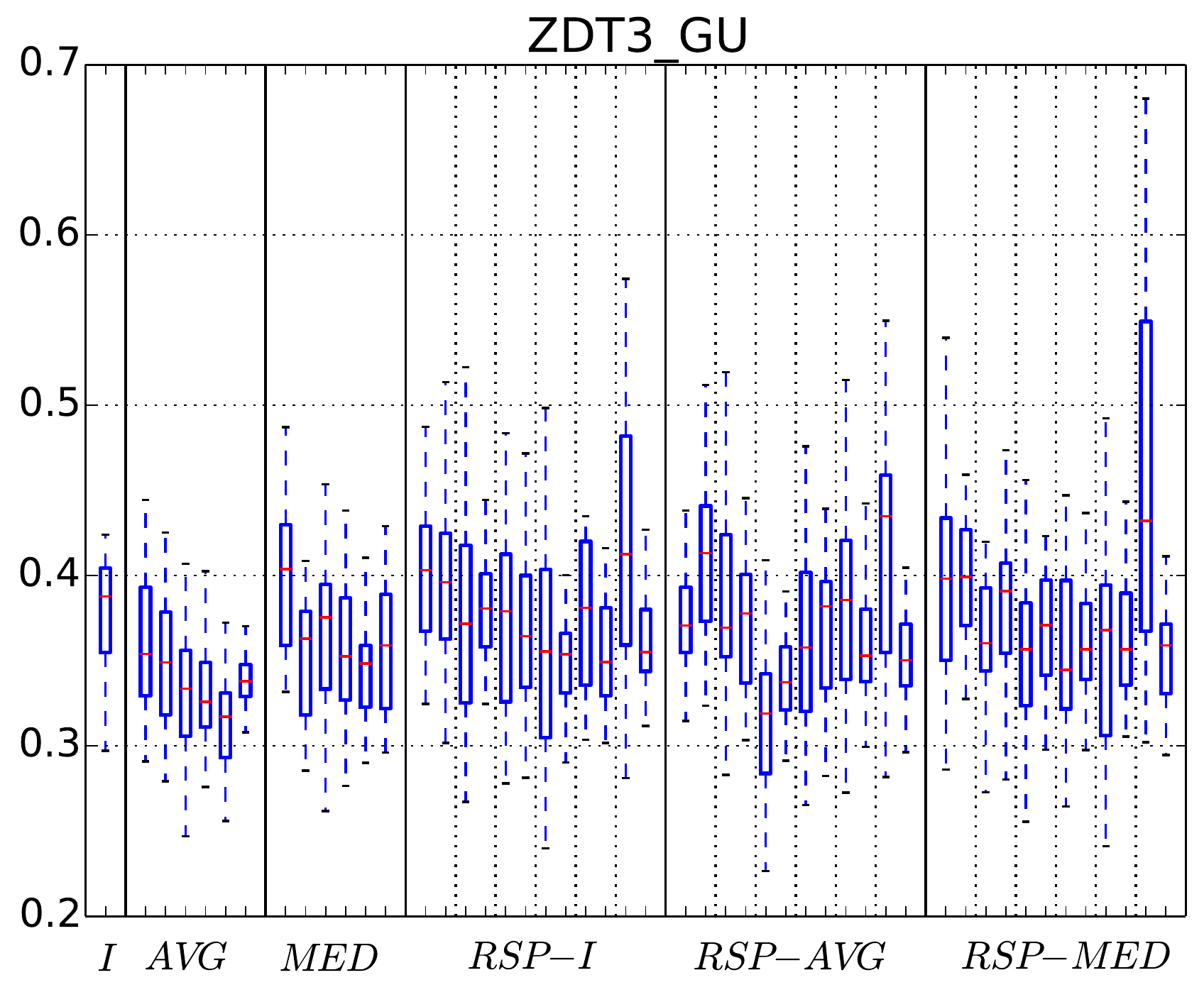}\\
}
\caption{Results for ZDT1 (4 top plots) and ZDT3 (4 bottom plots). See Section \ref{details-results}.}
\end{figure}

\begin{figure}[tb]
\hskip 0cm
\parbox{\textwidth}{
    \includegraphics[width=0.5\textwidth,height=0.15\textheight]{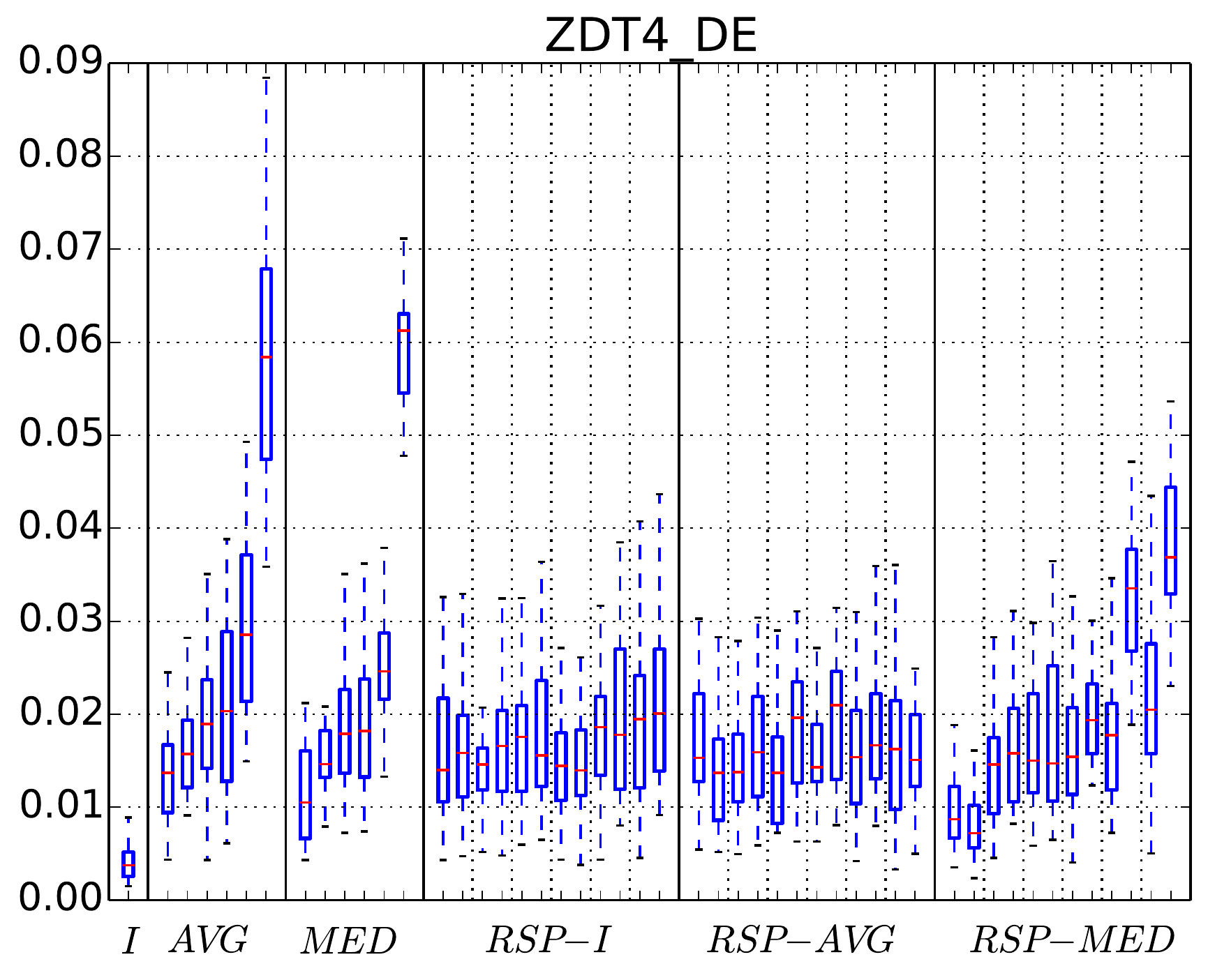}
\hskip 0cm
    \includegraphics[width=0.5\textwidth,height=0.15\textheight]{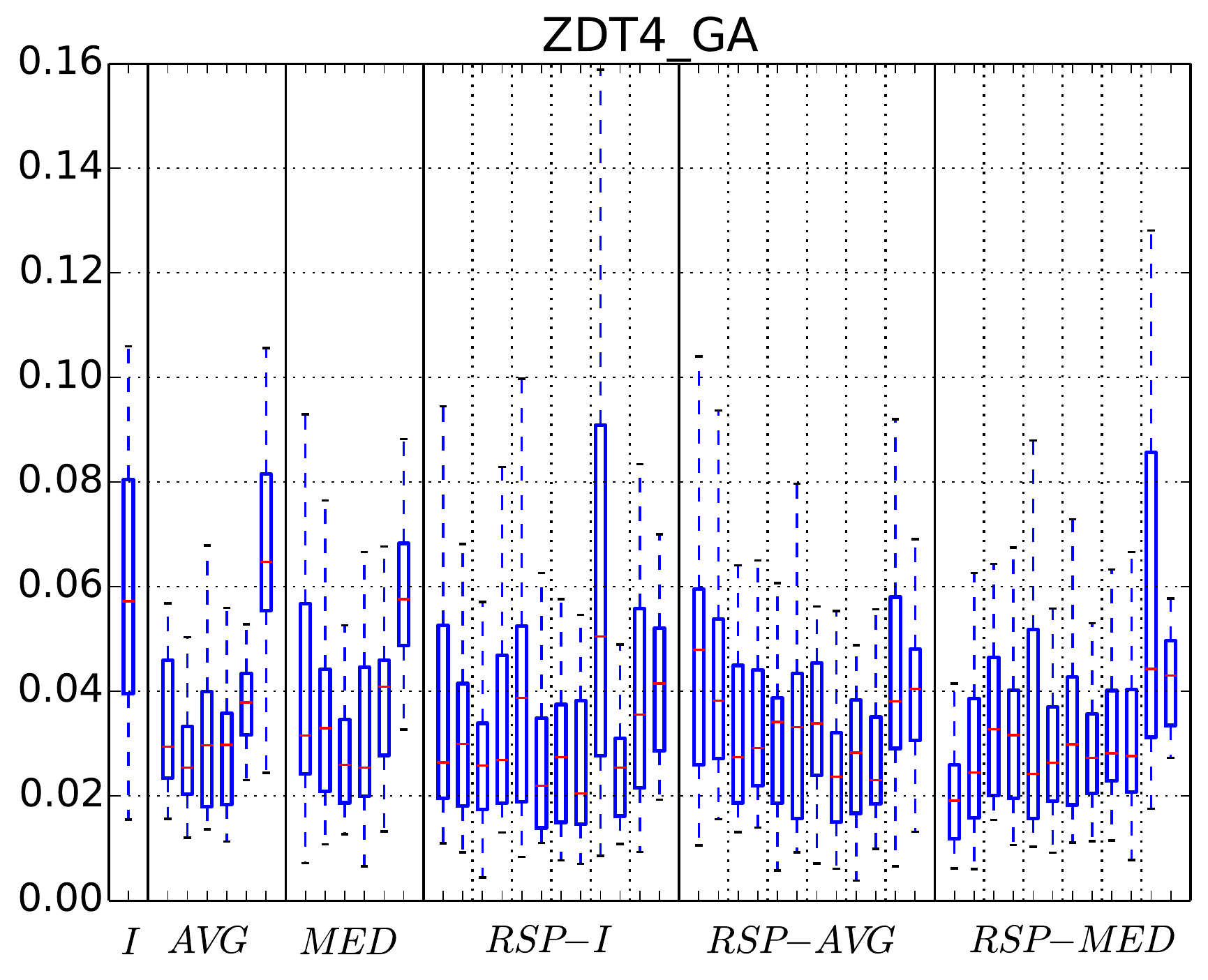}\\
    \includegraphics[width=0.5\textwidth,height=0.15\textheight]{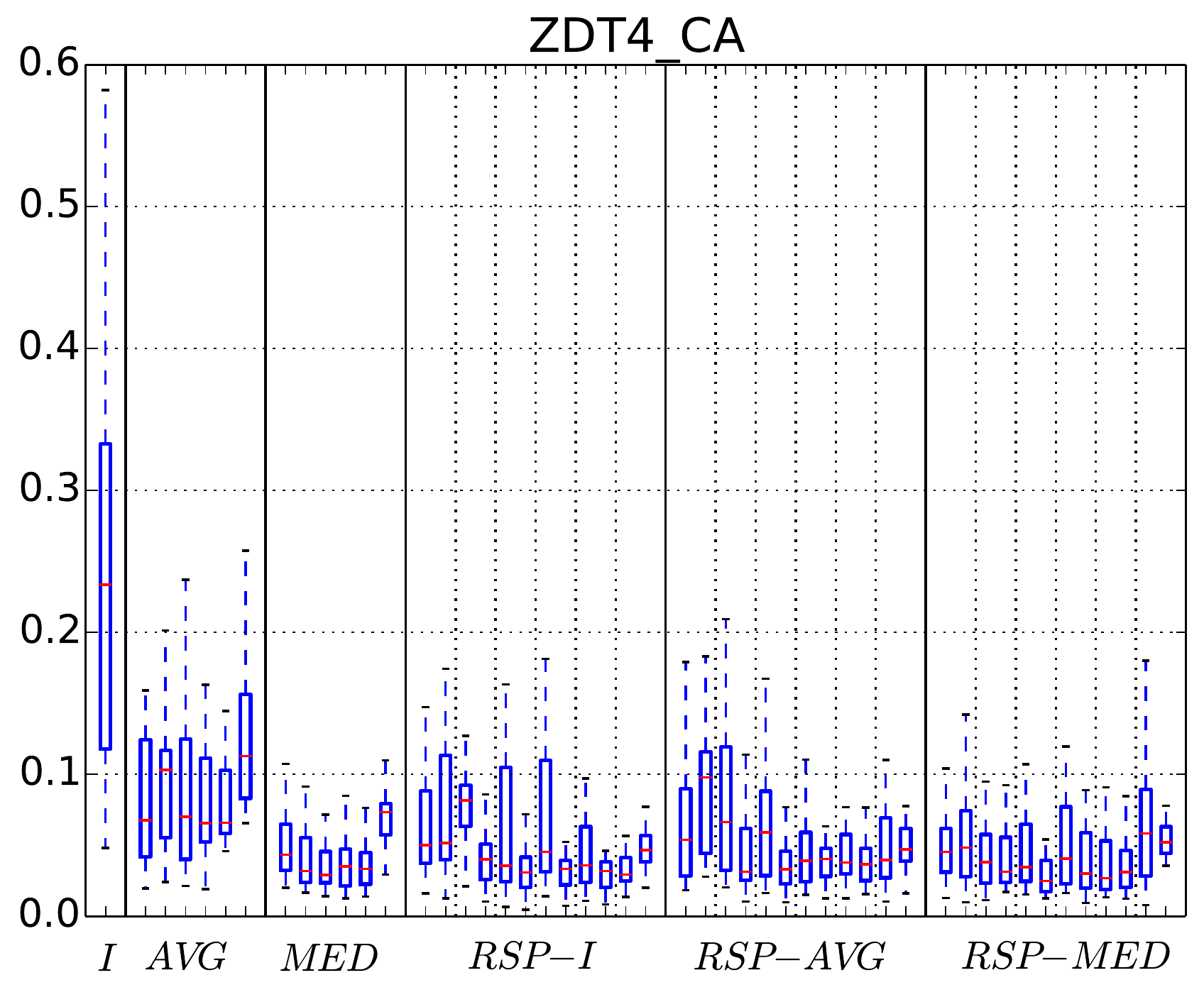}
\hskip 0cm
    \includegraphics[width=0.5\textwidth,height=0.15\textheight]{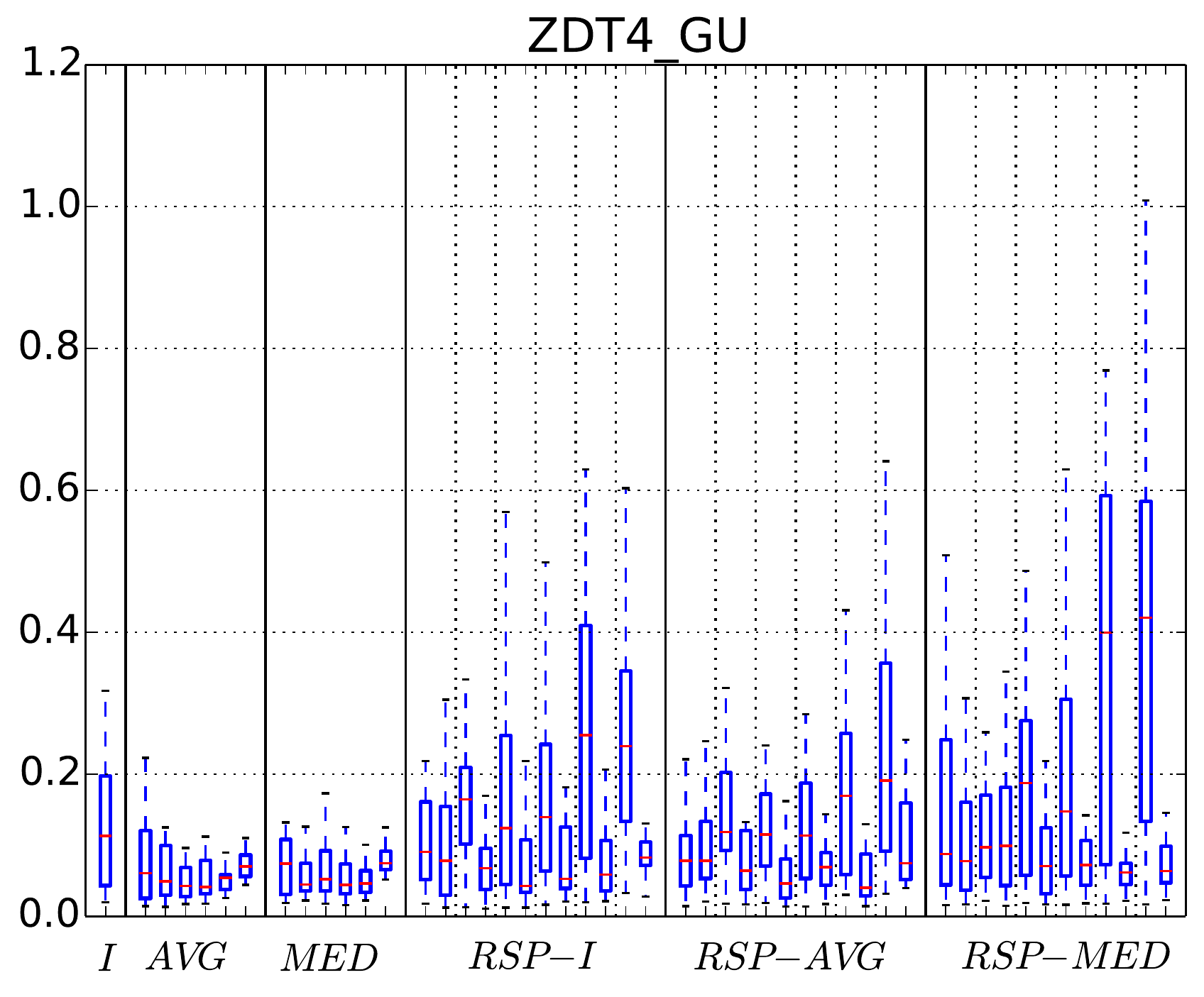}\\
    \includegraphics[width=0.5\textwidth,height=0.15\textheight]{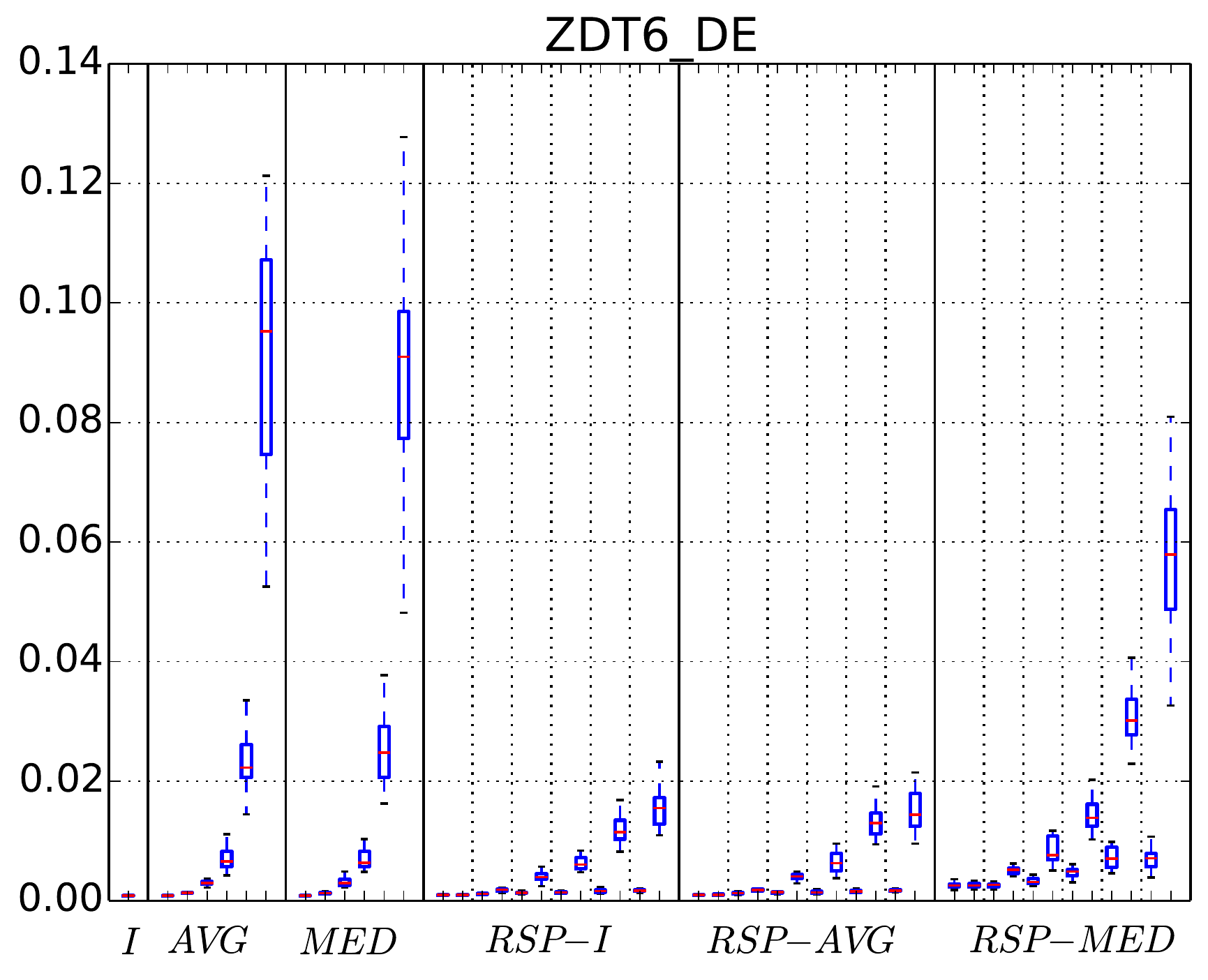}
\hskip 0cm
    \includegraphics[width=0.5\textwidth,height=0.15\textheight]{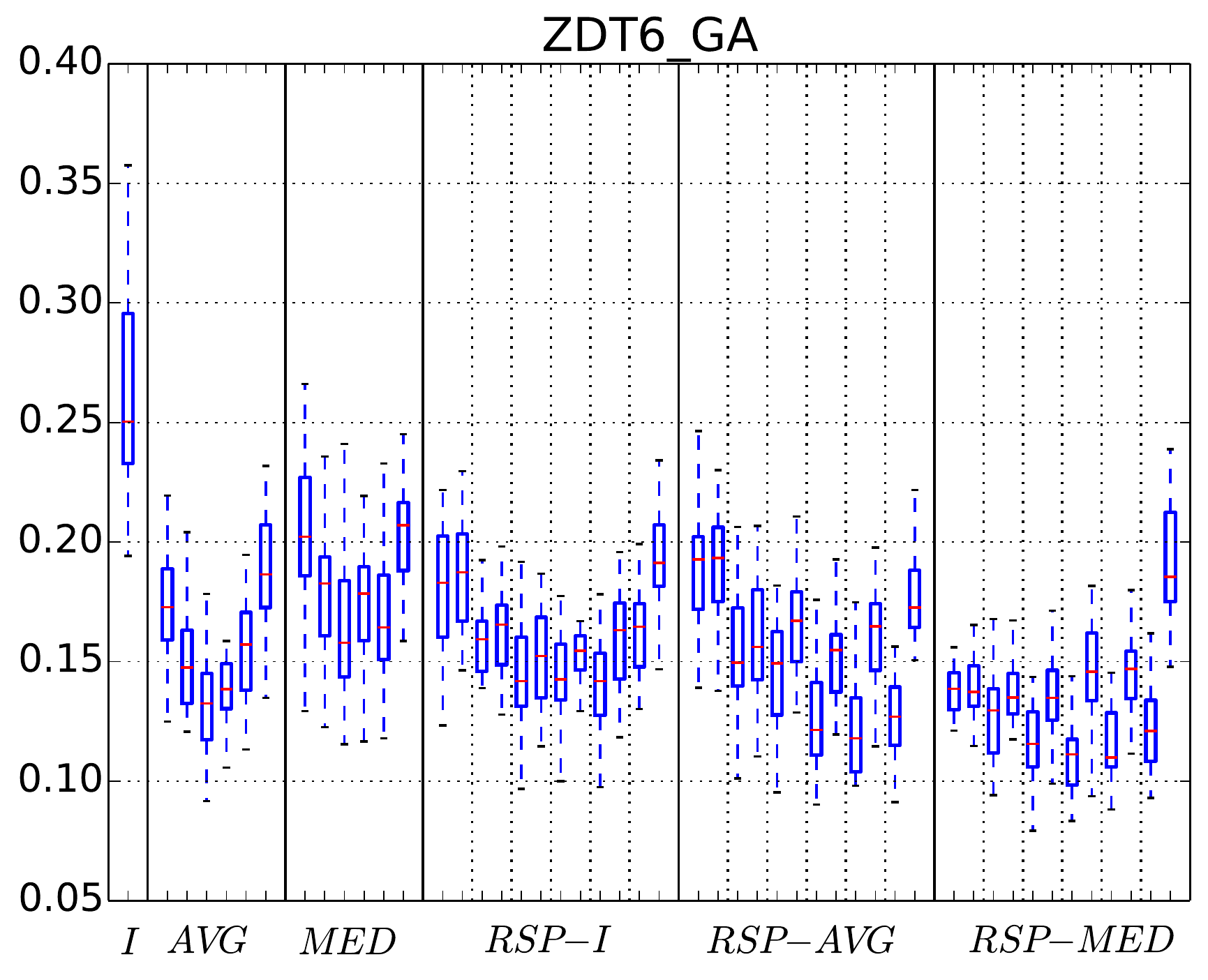}\\
    \includegraphics[width=0.5\textwidth,height=0.15\textheight]{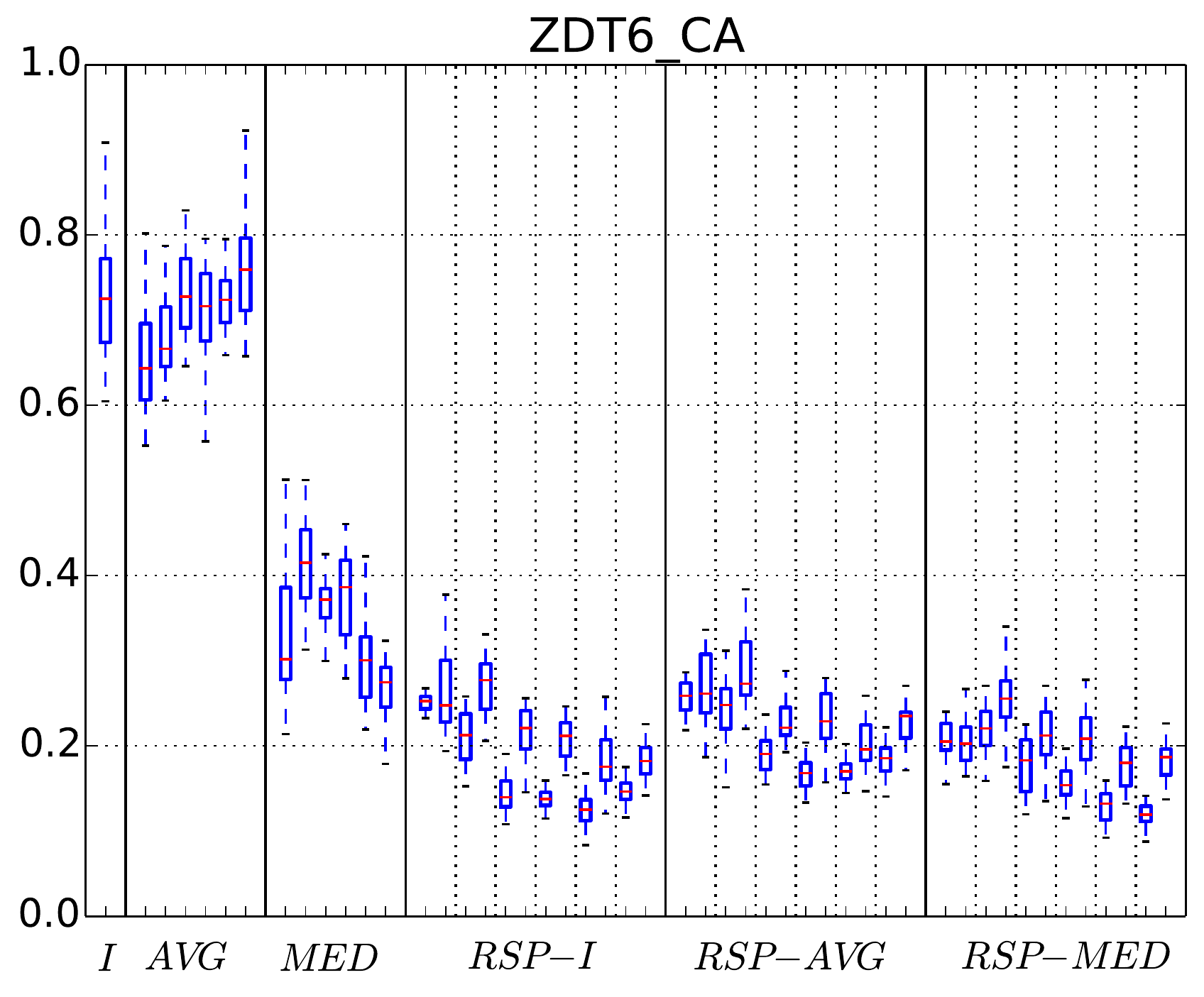}
\hskip 0cm
    \includegraphics[width=0.5\textwidth,height=0.15\textheight]{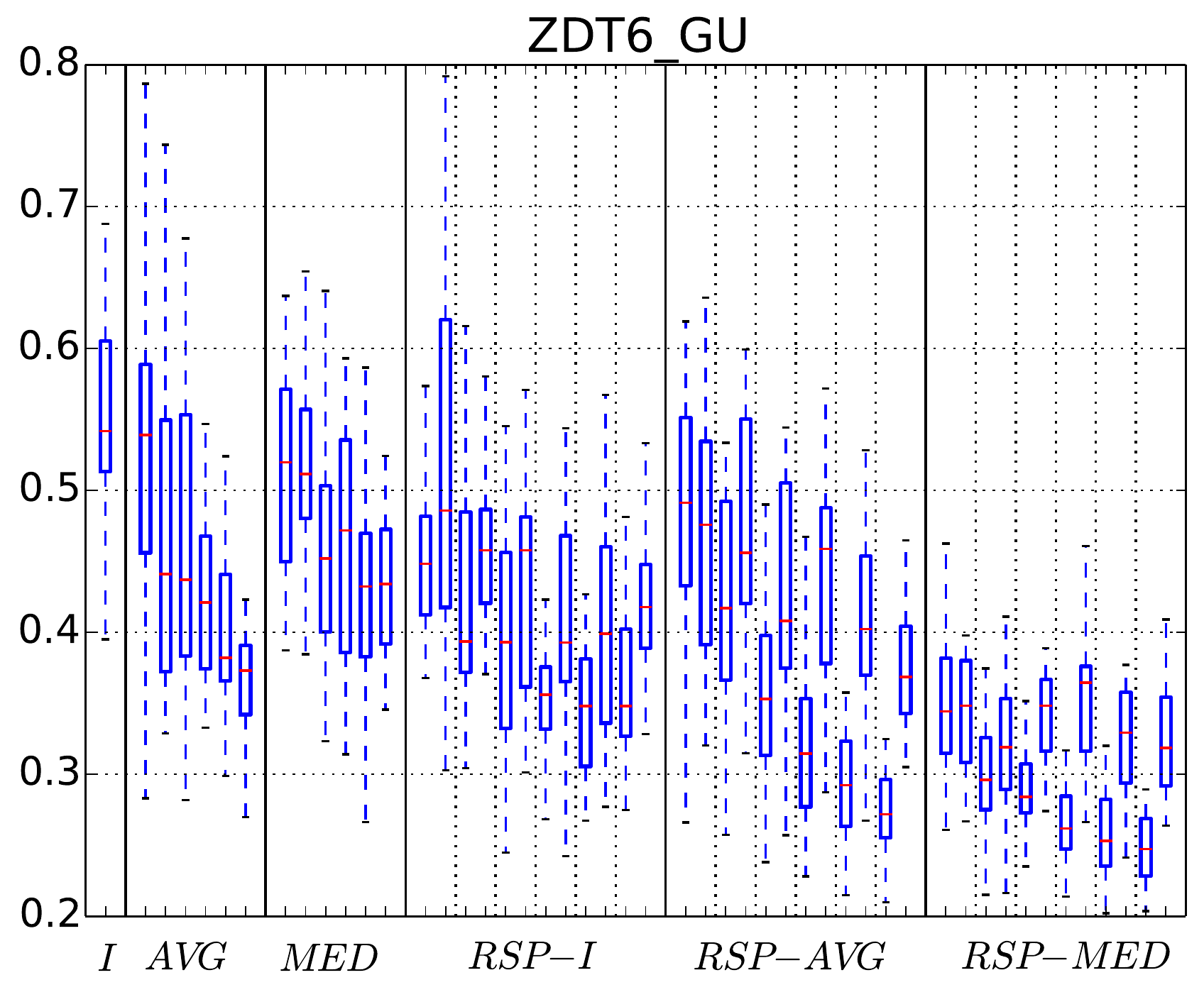}
}
\caption{Results for ZDT4 (4 top plots) and ZDT6 (4 bottom plots). See Section \ref{details-results}.}
\end{figure}



\subsection{Discussion}
First of all, in the {\bf deterministic case}, the results of implicit averaging assesses that the total budget of 100k samples is sufficient for NSGA-II to find a good approximation of the Pareto front. 
Furthermore, as expected, the performance of static resampling using an estimator degrades with the {\em Sampling Budget}, as more and more samples are wasted on  the (sometimes useless) estimation of the statistic. In the same situation, RSP is able to detect the low (!) uncertainty and to stop the race early, at least when using a {\em Confidence Level} of 25\%. A {\em Confidence Level} of 95\% can sometimes, on the other hand, lead to a similar degradation than in the static setting. The anomalies in that respect for RSP on ZDT3 (discontinuous front) for small {\em Sampling Budget} (5 and 10) might come from races that stop too early with all selected individuals in the same component of the front.



On the {\bf noisy instances}, implicit averaging does not perform very well compared to the other uncertainty handling approaches.  
Surprisingly, even if the medians are higher, the spread of the performances is not greater than the other approaches, excepted for ZDT4-CA.
It can be due to the fact that without any uncertainty handling approach, the probability that every individual of the population is good or bad is small and so, at the population level, the performance does not vary so much from one run to another.\\
\indent Beside, implicit averaging is comparable to AVG in case of Cauchy noise, for all functions but ZDT4: choosing by default the mean (a common choice) can lead to poor results when the distribution of the noise is unknown. 
Using RSP seems to mitigate this effect, probably because it uses the probability of survival instead of the estimator of the mean. 

Comparing, for each noisy function, the best configurations of RSP and static sampling leads to the following considerations: the results are statistically equivalent for all cases of noisy ZDT2 and ZDT4; 
RSP is significantly better (p-value $< 10^{-5}$) than static sampling in 5 cases (the 3 noisy ZDT6, and ZDT1 and 3 with Cauchy noise), is slightly worse (p-value in $[0.01,0.1]$) in 2 cases (both ZDT1 and 3 with Gaussian noise), and both approaches are equivalent (p-value $> 0.1$) on the remaining 2 cases (both ZDT1 and 3 with Gumbel noise). 
On ZDT1 and 3 with Gaussian noise, static sampling with averaging performs best: 
this is most probably related to the fact that AVG is based on the minimum-variance unbiased estimator, while $RSP_{AVG}$ uses it indirectly to estimate the probability of survival.


Regarding the choice of estimator, racing seems to decrease the impact of the average vs median issue. Indeed, when using static sampling, average performs slightly better than median for Gaussian and Gumbel noises, whereas median is consistently and significantly better when facing Cauchy noise. On the opposite, all 3 variants of RSP perform in general similarly over all problems. In particular, the no-estimator, $RSP_I$, performs as good as both others on most problems. 
This is good news, as it gives hope that the proposed racing approach might perform well with a lot of estimators, allowing the user to actually choose his favorite without having to care about the optimization algorithm in that respect.



\section{Conclusion and Perspective}
RSP is a general approach to uncertainty handling in existing EMOAs. It uses a ($\mu, \lambda$) Hoeffding race at a given confidence level, inspired by \cite{Heidrich-Igel-ICML09}, though applied directly on the selection probabilities of the individuals in the population. It is agnostic w.r.t. the selection method, and hence can accomodate any user preference that could be carried by the algorithm selection.

First experimental results within NSGA-II on noisy versions of ZDT benchmarks, indicate that this path is worth following for future research: RSP performs significantly better than implicit averaging or static sampling in many situations, and never performs significantly worse. It is less sensitive to the {\em Sampling Budget} parameter, especially for small (on zero) levels of noise, and surprisingly almost insensitive to the choice of the estimator. On the other hand, it is very sensitive to the {\em Confidence Level} of the races. However, these partial conclusions should be sustained by deeper analyses and validated by more experiments, with different levels of non-homogeneous noise, and other test functions from real-world problems.

The main perspectives for further work are to couple RSP with other EMOAs such as SPEA-2, IBEA and HYPE, in order to study the interaction between racing and the indicator function. Also, RSP should also be compared to more sophisticated uncertainty handling methods (see Section \ref{State-of-the-art}). 
It is also mandatory to test other estimators within RSP, as well as different noise models and different noise intensities. On the more fundamental side, it should be possible to better understand the intricate relationship between estimating the selection probability and directly estimating the objective values.

A longer term research track is to come up with some adaptive procedure to dynamically tune the {\em Sampling Budget} and, maybe more importantly, the {\em Confidence Level}. 
Indeed, it is clear from the present results that this latter parameter has a strong effect on the performance of the algorithm and should be fixed carefully. In the case where its optimal value varies over the decision space, only adaptive tuning can perform well on most functions.
To conclude, we feel that the use of RSP in EMOA is a promising avenue for taking into account the decision maker's preferences and increasing the reliability and robustness of the solutions in many real-world applications.

%
%
%
%

\bibliographystyle{splncs03}
\bibliography{ppsn}

\begin{thebibliography}{10}
\providecommand{\url}[1]{\texttt{#1}}
\providecommand{\urlprefix}{URL }

\bibitem{basseurZitzler:EvoApps2006}
Basseur, M., Zitzler, E.: {A Preliminary Study on Handling Uncertainty in
  Indicator-Based Multiobjective Optimization}. In: {F. Rothlauf et al.} (ed.)
  Proc. EvoWorkshops'06. pp. 727--739. LNCS 3907, Springer Verlag (2006)

\bibitem{BoonmaSuzuki:ICTAI2009}
Boonma, P., Suzuki, J.: {A Confidence-based Dominance Operator in Evolutionary
  Algorithms for Noisy Multiobjective Optimization Problems}. In: Proc.
  ICTAI'09. IEEE Press (2009)

\bibitem{DebBook2001}
Deb, K.: Multi-Objective Optimization Using Evolutionary Algorithms. John Wiley
  \& Sons (2001)

\bibitem{Eskandari:CEC07}
Eskandari, H., Geiger, C.D., Bird, R.: {Handling uncertainty in evolutionary
  multiobjective optimization: SPGA}. In: CEC'07. pp. 4130--4137. IEEE Press
  (2007)

\bibitem{Fieldsend:CEC2005}
Fieldsend, J., Everson, R.: {Multi-Objective Optimisation in the Presence of
  Uncertainty}. In: CEC'05. pp. 243--250. IEEE Press (2005)

\bibitem{Heidrich-Igel-ICML09}
Heidrich-Meisner, V., Igel, C.: {Hoeffding and Bernstein Races for Selecting
  Policies in Evolutionary Direct Policy Search}. In: {A.P. Danyluk et al.}
  (ed.) Proc. ICML'09. ACM Intl Conf. Proc. Series, vol. 382, p.~51 (2009)

\bibitem{Hughes:EMO2001}
Hughes, E.J.: {Evolutionary Multiobjective Ranking with Uncertainty and Noise}.
  In: {E. Zitzler et al.} (ed.) EMO'01. pp. 329--343. LNCS 1993, Springer
  Verlag (2001)

\bibitem{Phan:2012}
Phan, D.H., Suzuki, J.: {A Non-parametric Statistical Dominance Operator for
  Noisy Multiobjective Optimization}. In: {Lam Thu Bui et al.} (ed.) Proc.
  SEAL'12. pp. 42--51. Springer-Verlag (2012)

\bibitem{masterBranke2009}
Siegmund, F.: Sequential Sampling in Noisy Multi-Objective Evolutionary
  Optimization. Master's thesis, School of Humanities and Informatics, Sk\"ovde
  (2009)

\bibitem{Teich:EMO2001}
Teich, J.: {Pareto-Front Exploration with Uncertain Objectives}. In: {E.
  Zitzler et al.} (ed.) Proc. EMO'01. pp. 314--328. LNCS 1993, Springer Verlag
  (2001)

\bibitem{trautmanNaujoks:CEC2009}
Trautmann, H., Mehnen, J., Naujoks, B.: {Pareto-Dominance in Noisy
  Environments}. In: Tyrrell, A. (ed.) Proc. CEC'09. pp. 3119--3126. IEEE Press
  (2009)

\bibitem{vossIgel:PPSN2010}
Vo\ss, T., Trautmann, H., Igel, C.: {New Uncertainty Handling Strategies in
  Multi-objective Evolutionary Optimization}. In: {R. Schaefer et al.} (ed.)
  Proc. PPSN XI. pp. 260--269. LNCS 6239, Springer Verlag (2010)

\bibitem{surveyZhou2011}
Zhou, A., Qu, B.Y., Li, H., Zhao, S.Z., Suganthan, P.N., Zhang, Q.:
  Multiobjective evolutionary algorithms: A survey of the state of the art.
  Swarm and Evolutionary Computation  1(1),  32 -- 49 (2011)

\end{thebibliography}

\end{document}